\documentclass[11pt, logo, onecolumn, copyright, colorlinks=true, allcolors=blue]{nvidiatechreport}
\usepackage[round]{natbib}
\usepackage{listings}
\usepackage{subcaption}
\usepackage{xcolor}
\usepackage{xspace}
\usepackage[varqu,varl]{inconsolata}   
\usepackage[skins,breakable,listings]{tcolorbox}
\usepackage{placeins}
\usepackage{tikz}
\usetikzlibrary{calc}
\usepackage{xurl}
\usepackage{hyperref}
\usepackage{booktabs}
\usepackage{graphicx}
\usepackage[font=small,labelfont=bf]{caption}

\definecolor{nvgreen}{RGB}{118,185,0}      
\definecolor{nvamber}{HTML}{E8A33D}
\definecolor{nvblue}{HTML}{2E7DB8}
\definecolor{nvpurple}{HTML}{8B5CF6}
\definecolor{nvlightgray}{HTML}{D1D5DB}
\definecolor{harnessgray}{HTML}{6B7280}
\definecolor{diagink}{HTML}{1A1A1A}
\definecolor{diagmute}{HTML}{6B7280}
\definecolor{diagline}{HTML}{E8EAED}
\definecolor{diagstroke}{HTML}{CFD3D8}
\definecolor{diagtask}{HTML}{E5E7EB}
\definecolor{diagamberfill}{HTML}{FBE7C2}
\definecolor{diaggreenfill}{HTML}{E4F2C4}
\definecolor{diagarch}{HTML}{9AA1AB}

\definecolor{kwgreen}{RGB}{84,128,18}
\definecolor{builtinblue}{RGB}{41,103,160}
\definecolor{strteal}{RGB}{17,133,128}
\definecolor{commentgreen}{RGB}{70,120,70}
\definecolor{codebg}{RGB}{248,249,246}
\definecolor{codeframe}{RGB}{224,227,220}
\definecolor{codebackground}{HTML}{F6F8FA}

\newtcolorbox{codebox}{
  enhanced,
  colback=codebg,
  colframe=codeframe,
  boxrule=0.4pt,
  arc=2.5pt,
  leftrule=0pt,
  borderline west={2.2pt}{0pt}{nvgreen},
  boxsep=2pt, left=10pt, right=8pt, top=5pt, bottom=5pt,
  before skip=6pt, after skip=6pt,
}

\newtcolorbox{findingsbox}{
  enhanced,
  colback=nvgreen!8,
  colframe=nvgreen!95,
  boxrule=0pt,
  leftrule=3pt,
  arc=0pt,
  left=8pt, right=8pt, top=4pt, bottom=4pt,
  before skip=0.6em, after skip=0.8em,
}

\newtcbox{\inlinecode}[1][]{
  on line, boxrule=0pt, colback=codebackground, coltext=black, arc=2pt,
  before upper={\strut\ttfamily\small}, boxsep=0pt, left=2pt, right=2pt, top=0pt, bottom=0pt, #1
}

\newcommand{\nooa}{\textsc{NOOA}\xspace}
\newcommand{\authorsep}{\hspace{0.8em}}
\newcommand{\authorsepp}{\hspace{1.5em}}

\newcommand{\legline}[2][]{\tikz[baseline=-0.6ex]\draw[#2, line width=1.4pt, #1] (0,0) -- (0.6,0);}
\newcommand{\legdash}[1]{\legline[dash pattern=on 2.5pt off 1.5pt]{#1}}
\newcommand{\legbox}[1]{\tikz[baseline=-0.6ex]\fill[#1] (0,-0.1) rectangle (0.22,0.12);}
\newcommand{\legitem}[2]{\mbox{#1\,#2}\hspace{1.1em}}
\newcommand{\figlegend}[1]{\par\centering{\footnotesize\sffamily #1}\par\smallskip}
\newcommand{\paneltitle}[2]{{\small\sffamily\textbf{#1}}\\[-1pt]{\scriptsize\sffamily\color{diagmute}#2}\par\vspace{2pt}}

\definecolor{darkblue}{rgb}{0, 0, 0.5}

\hypersetup{
    colorlinks=true,        
    linkcolor=darkblue,   
    citecolor=darkblue,   
    filecolor=darkblue,   
    urlcolor=darkblue     
}

\title{Letting Models Manage Their Own Context: Where We Stand}

\author{%
\begin{minipage}{\linewidth}
\centering
Alessio Devoto\authorsepp
Avinash Vem\authorsepp
Tara Safavi\par\vspace{0.25em}
Sanjay Kariyappa\authorsep
Yu-Hsuan Tseng\authorsep
Paul Furgale\authorsep
Ricardo Silveira Cabral
\par\vspace{1.2em}
\end{minipage}
}

\begin{document}

\begin{abstract}
Agent harnesses today manage context on the model's behalf: when the window fills, the harness triggers a compaction that summarizes the history. We study the alternative, \emph{model-driven} context management, in which the model itself decides when and what to compact through a context API. We gave models control over their own context in \nooa, NVIDIA's open source agent harness, and compared the two strategies on document research and machine learning engineering across context budgets from 8K to 512K tokens. Models that use the context API effectively reach higher accuracy, often with fewer input tokens, especially at tight budgets: \textbf{GPT 6.1 nearly triples its score at a 32K budget} on OfficeQA Pro v2 (21\%~$\rightarrow$~56\%), and \textbf{Kimi gains with up to 41\% fewer input tokens}. Models differ widely, however, in how often they use the API, how aggressively they compact, and when they start compacting, and prompt changes shift that timing. Finally, we show that \textbf{inference engines can make model-driven edits cheaper}: by reusing the KV cache of the unchanged suffix after an edit, \textbf{we cut prefill compute by up to 15\% per request with no measurable change in accuracy}. Together, these results point to a new co-design frontier: models trained to manage their own context, served by inference engines that support KV cache editing.
\end{abstract}

\maketitle

\section{Introduction}
\label{sec:intro}

When we released \nooa~\citep{nooa}, NVIDIA's open source agent harness, we built it around six core ideas. One was to give the model direct control over the harness through APIs, including context compaction. The motivation was simple: an agent should be able to decide which parts of its history to keep in context, which to summarize, and which to retrieve when needed~\citep{sixcapabilities}.

Interest in model-driven context management has grown since. Research such as SideQuest~\citep{kariyappa2026sidequest} and Context Language Models~\citep{shao2026clm} explores the idea directly, and agent harnesses are starting to let models manage their own context too. Codex, for example, has introduced an experimental context-management mode~\citep{codexrelease} that combines history and notes with a \inlinecode{new\_context} tool for starting a fresh context window.

This raises practical questions: \textbf{Can current models manage their own context effectively? Does it improve task performance? And can inference engines turn model-managed context into cheaper inference?}

We compared two context management strategies on document research and machine learning engineering tasks. In \emph{harness-driven} context management, the default in today's harnesses, the harness triggers compaction when the context window fills. In \emph{model-driven} context management, the model decides when and what to compact. Four takeaways stand out:

\begin{findingsbox}
\begin{itemize}[leftmargin=1.2em, itemsep=0.45em, topsep=0.2em]
  \item \textbf{Model-driven context management can push the accuracy-efficiency frontier.} Models that use \nooa's context API effectively reach higher accuracy, often with fewer input tokens, especially at tight context budgets: GPT 6.1 Sol nearly triples its score at a 32K context budget on OfficeQA Pro v2 (21\%~$\rightarrow$~56\%; Section~\ref{sec:promise}).
  \item \textbf{Models differ in how they use the API and when they compact.} Some barely use it, some compact in large, aggressive steps, and some compact earlier than they need to. Prompt changes shift when they compact (Sections~\ref{sec:models-differ} and~\ref{sec:anxiety}).
  \item \textbf{Inference engines can make model-driven edits cheaper.} Today's engines reuse the KV cache only up to the first change in the prompt~\citep{vllmprefixcaching}, so every context edit forces the engine to recompute everything after it, including text that hasn't changed. By also reusing the KV cache of that unchanged suffix, we cut prefill compute by up to 15\% per request with no measurable change in accuracy (Section~\ref{sec:cheaper}).
  \item \textbf{This is a new frontier for co-design.} Realizing the full promise of model-driven context management will require models trained to manage their own context and inference engines built to serve it efficiently at scale (Section~\ref{sec:next}).
\end{itemize}
\end{findingsbox}
\section{From Harness-Driven to Model-Driven Context Management}
\label{sec:modes}

\begin{figure}[tbp]
  \centering
\begin{tikzpicture}[x=\linewidth/720, y=-\linewidth/720, every node/.style={inner sep=0pt, outer sep=0pt}]
  \tikzset{
    b/.style={draw=diagstroke, line width=0.5pt, rounded corners=1.5pt},
    task/.style={b, fill=diagtask},
    ev/.style={b, fill=white},
    sel/.style={b, fill=diagamberfill, draw=nvamber},
    sum/.style={b, fill=diaggreenfill, draw=nvgreen},
    arch/.style={b, fill=white, draw=diagarch, dash pattern=on 2pt off 1.5pt},
    hd/.style={anchor=base west, font=\sffamily\bfseries\small, text=diagink},
    sub/.style={anchor=base west, font=\sffamily\scriptsize, text=diagmute},
    row/.style={anchor=base west, font=\sffamily\scriptsize, text=diagmute},
    note/.style={anchor=base west, font=\sffamily\tiny, text=diagmute},
    notec/.style={note, anchor=base},
  }
  \newcommand{\blk}[6]{\draw[#5] (#1,#2) rectangle ++(#3,#4);
    \node[font=\sffamily\tiny, text=diagink] at ({#1+#3/2},{#2+#4/2}) {#6};}

  \node[hd]  at (0,16) {Harness-driven};
  \node[sub] at (0,34) {The harness decides when to compact};
  \node[row] at (0,66) {Before};
  \blk{70}{48}{58}{34}{task}{Task}
  \foreach \i [evaluate=\i as \n using int(\i+1)] in {0,...,11}
    {\blk{134+\i*38}{48}{34}{34}{ev}{e\n}}
  \draw[diagmute, line width=0.8pt, dash pattern=on 2pt off 1.5pt] (596,40) -- (596,92);
  \node[note] at (602,62) {usage crosses 80\%};
  \node[note] at (602,77) {$\rightarrow$ harness asks a model};
  \node[note] at (602,91) {to summarize};

  \node[row] at (0,126) {After};
  \blk{70}{108}{58}{34}{task}{Task}
  \blk{134}{108}{150}{34}{sum}{Summary of e1--e2}
  \foreach \i [evaluate=\i as \n using int(\i+3)] in {0,...,9}
    {\blk{290+\i*38}{108}{34}{34}{ev}{e\n}}
  \node[note] at (134,158) {Initial instructions and the 10 most recent events are kept verbatim.};

  \draw[diagline] (0,172) -- (720,172);

  \node[hd]  at (0,188) {Model-driven};
  \node[sub] at (0,206) {The model decides when and what to compact};
  \node[row] at (0,238) {Before};
  \blk{70}{220}{58}{34}{task}{Task}
  \foreach \i [evaluate=\i as \n using int(\i+1)] in {0,...,11}{
    \ifnum\i<2 \def\st{ev}\else\ifnum\i>8 \def\st{ev}\else\def\st{sel}\fi\fi
    \blk{134+\i*38}{220}{34}{34}{\st}{e\n}}
  \draw[nvamber, line width=0.8pt] (210,262) -- ++(0,6) -- ++(262,0) -- ++(0,-6);
  \node[notec] at (343,282) {model picks e3--e9 and writes a summary};
  \node[note] at (602,240) {usage indicator: 85\%};

  \node[row] at (0,328) {After};
  \blk{70}{310}{58}{34}{task}{Task}
  \blk{134}{310}{34}{34}{ev}{e1}
  \blk{172}{310}{34}{34}{ev}{e2}
  \blk{210}{310}{150}{34}{sum}{Summary of e3--e9}
  \foreach \i [evaluate=\i as \n using int(\i+10)] in {0,...,2}
    {\blk{366+\i*38}{310}{34}{34}{ev}{e\n}}
  \blk{494}{310}{124}{34}{arch}{archive (by tag)}
\end{tikzpicture}
  \caption{In harness-driven context management (top), the harness triggers compaction when context usage reaches a threshold. In model-driven context management (bottom), the model edits its own context along the way.}
  \label{fig:modes}
\end{figure}
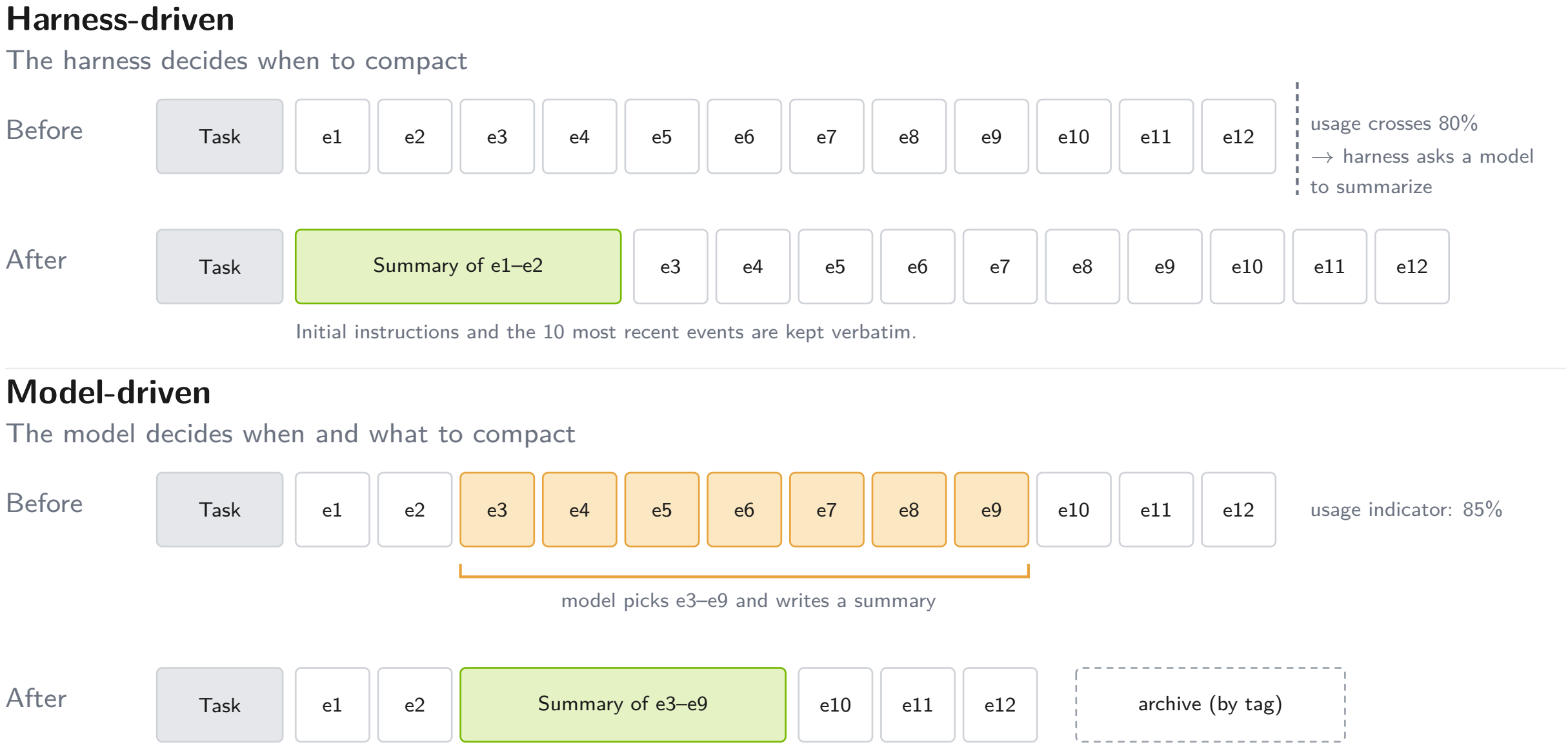

As an agent works on a task, its history of tool calls, outputs, and model turns keeps growing, making it harder for the model to find and use relevant information (context rot~\citep{contextrot}) and eventually exceeding the context window.

Most harnesses handle this automatically: when context usage crosses a threshold, the harness summarizes the history (Figure~\ref{fig:modes}, top), a process commonly referred to as compaction. We call this \textbf{harness-driven context management}. The harness' heuristic policy decides when compaction happens and which parts of the history it covers, while the model only generates the summary. While simple, this approach can be lossy and lead to loss of key task-relevant information in the model's context.

\textbf{Model-driven context management} takes a different approach and lets the model decide when and what to compact (Figure~\ref{fig:modes}, bottom). In \nooa, the model can inspect its history and replace a range of events with a summary through its context API (the \inlinecode{collapse} call). Compacted events are archived, so the model can retrieve the originals by tag.

Listing~\ref{lst:collapse} shows the two operations it needs: collapsing a range of events into a summary, and retrieving an archived event by its original tag.

\begin{figure}[h]
\begin{codebox}
\begin{lstlisting}
# Replace events 3..17 with a summary; returns the summary tag "3..17"
summary_tag = self.events.collapse(
    "3", "17", summary_text="Explored the repo layout; tests live in tests/unit."
)

# The originals are archived, not deleted: retrieve any event by its original tag
original = self.events["9"]
\end{lstlisting}
\end{codebox}
\vspace{-0.6em}
\captionsetup{name=Listing}
\caption{Collapsing events and retrieving an archived event by its original tag through \nooa's context API.}
\label{lst:collapse}
\end{figure}

\section{Experimental Setup}
\label{sec:setup}

To test model-driven context management across different kinds of work, we use two benchmarks.

\textbf{OfficeQA Pro v2}~\citep{officeqaprov2} contains 90 questions about U.S.\ federal receipts and expenditures. The agent answers them by searching and reading a large pre-parsed document corpus. Reading documents fills the context quickly, which makes this a good setting for studying how agents retain evidence across a research task.

\textbf{MLE-bench Lite}~\citep{chan2024mlebench, mlebenchlite} contains 22 Kaggle competitions, where the agent has to inspect datasets, write and debug code, evaluate results, and submit predictions. The agent has to track experiments, implementation choices, and execution feedback as it iterates.

We compare two context management strategies:
\begin{itemize}[leftmargin=1.5em, itemsep=0.2em]
  \item \textbf{Harness-driven:} at 80\% of the context budget, the harness summarizes the earlier history, preserving the initial instructions and the 10 most recent events.
  \item \textbf{Model-driven:} the model receives \nooa's context API, a live context-usage indicator, an instruction to consider compacting only once usage reaches 85\%, and a \emph{context-pressure warning} added to the request when the context approaches its limit.
\end{itemize}

We sweep context budgets from 8K to 512K tokens. Shrinking the budget raises \emph{context pressure}, the ratio of what the task needs to what the window holds, and simulates the effect of much longer tasks. Each budget is a hard limit on request size. If a request exceeds it, a fallback removes the oldest events and retries; after three failures, the run counts as a wrong answer.

We evaluate four models: We evaluate four models:
GPT-6.1 Sol~\citep{openai2026gpt61sol},
Kimi K3~\citep{kimik3},
DeepSeek V4 Pro~\citep{deepseek2026v4}, and
Claude Opus 5.5~\citep{anthropic2026opus55}.

\section{The Promise of Model-Driven Context Management}
\label{sec:promise}

\begin{figure}[tbp]
  \centering
  \figlegend{\legitem{\legdash{harnessgray}}{harness-driven}\legitem{\legline{nvgreen}}{model-driven}}
    \includegraphics[width=0.49\linewidth]{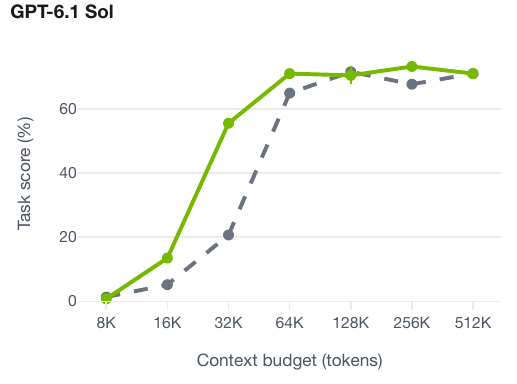}\hfill
    \includegraphics[width=0.49\linewidth]{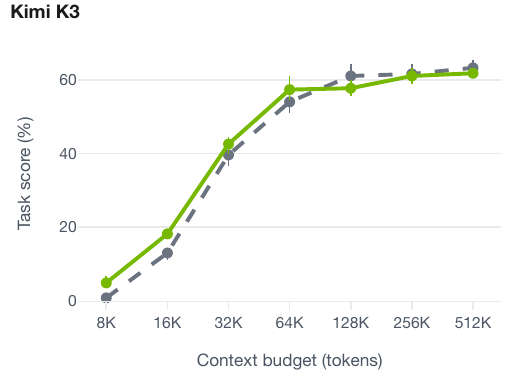}\\[0.4em]
    \includegraphics[width=0.49\linewidth]{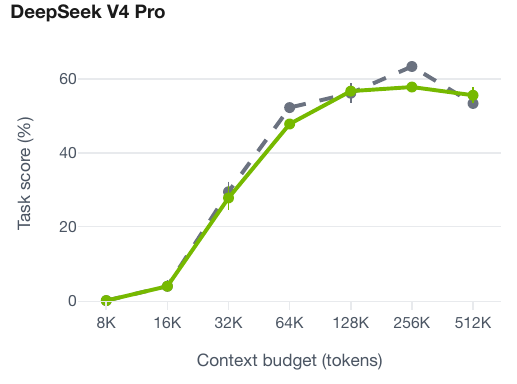}\hfill
    \includegraphics[width=0.49\linewidth]{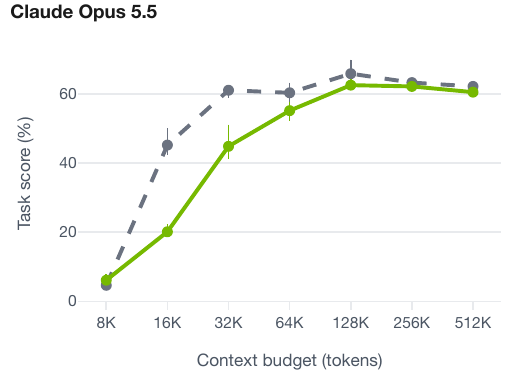}
  \caption{\textbf{Score vs.\ context budget, OfficeQA Pro v2}, harness-driven vs.\ model-driven, by model. Score is passed tasks over all 90 trials, including errors. Points are equal-weight replicate means; bars span replicate min--max.}
  \label{fig:score-budget}
\end{figure}

In our experiments, we found that models that understand and use \nooa's context API effectively can reach substantially higher accuracy at the same context budget compared to harness-driven compaction, without extra training. GPT-6.1 Sol shows the clearest benefit, improving accuracy by nearly $3\times$ over harness-driven context management at a 32K context budget on OfficeQA Pro v2 (Figure~\ref{fig:score-budget}). However, we found that gains are not consistent across all models: DeepSeek V4 Pro shows less improvement, and Opus 5.5 rarely uses the context API.

\begin{figure}[tbp]
  \centering
  \figlegend{\legitem{\legdash{harnessgray}}{harness-driven}\legitem{\legline{nvgreen}}{model-driven}}
    \includegraphics[width=0.49\linewidth]{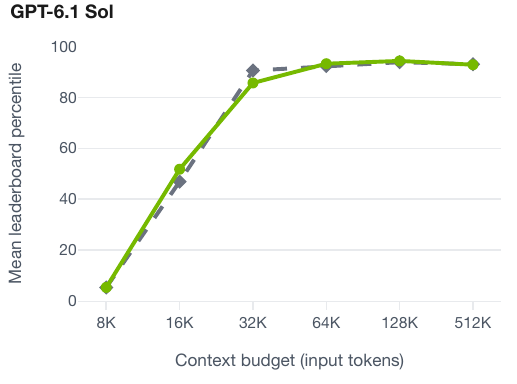}\hfill
    \includegraphics[width=0.49\linewidth]{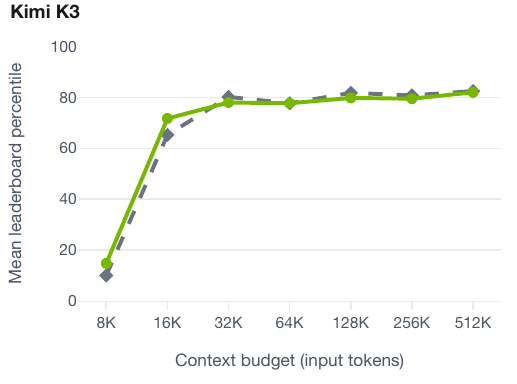}\\[0.4em]
    \includegraphics[width=0.49\linewidth]{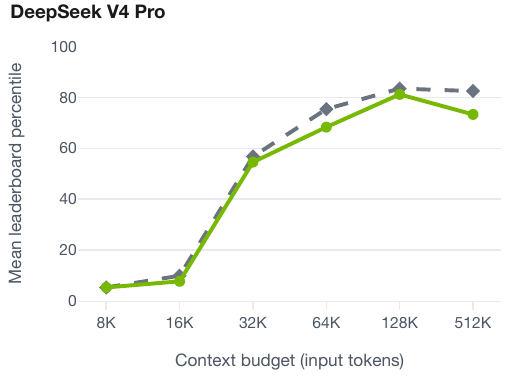}\hfill
    \includegraphics[width=0.49\linewidth]{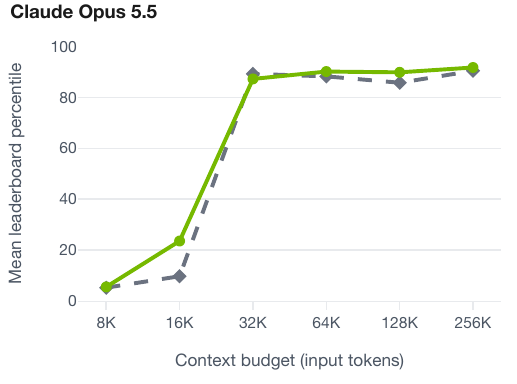}
  \caption{\textbf{Score vs.\ context budget, MLE-bench Lite}, harness-driven vs.\ model-driven, by model. Mean over 22 tasks. Scores are leaderboard percentile (missing scores count as zero). Each point averages the verified runs at that budget.}
  \label{fig:score-budget-mle}
\end{figure}

The strongest gains are at tight budgets, where the agent has to manage its history repeatedly to make progress, and they shrink as the budget grows. GPT and Kimi are ahead of the harness baseline at small budgets, while Opus trails it until the budget is large enough that the strategies converge. On MLE-bench Lite the effect is much smaller and limited to the tightest budgets (Figure~\ref{fig:score-budget-mle}), probably because its tasks need far less context.

This corroborates recent work~\citep{shao2026clm,fan2026harness,kontonis2026memento}, which finds that context management helps most when the window is tight relative to the task length.

The differences between models raise a further question: are they making different context-management decisions, or failing to use the capability at all? To find out, we looked at how each model used the API.

\section{Some Models Get It More Than Others}
\label{sec:models-differ}

\begin{figure}[tbp]
  \centering
  \includegraphics[width=0.62\linewidth]{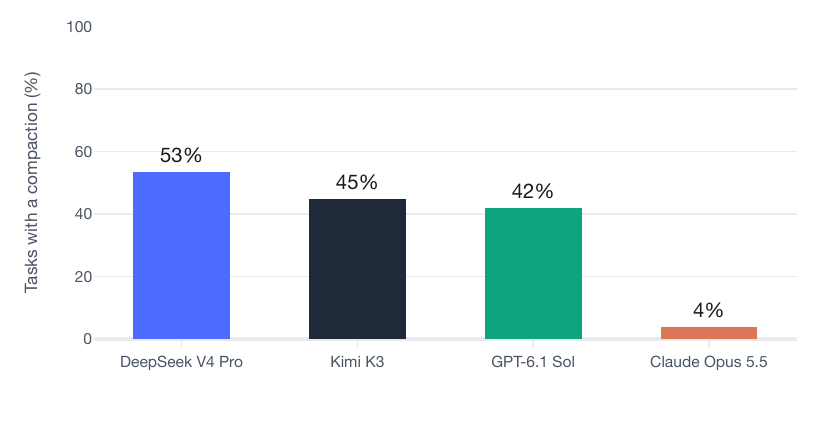}
  \caption{\textbf{Models differ in how often they use the context API.} Share of tasks with at least one compaction call, model-driven runs, all budgets.}
  \label{fig:adoption}
\end{figure}

All models receive the same context API and instructions, yet they use them very differently. GPT, Kimi, and DeepSeek call it in 42--53\% of tasks. Opus rarely reaches for it: it edits its context in only 4\% of tasks (Figure~\ref{fig:adoption}).

\begin{figure}[tbp]
  \centering
  \begin{subfigure}[t]{0.48\linewidth}
    \centering
    \paneltitle{Compression factor per compaction}{how many times smaller the request gets, log scale}
    \includegraphics[width=\linewidth]{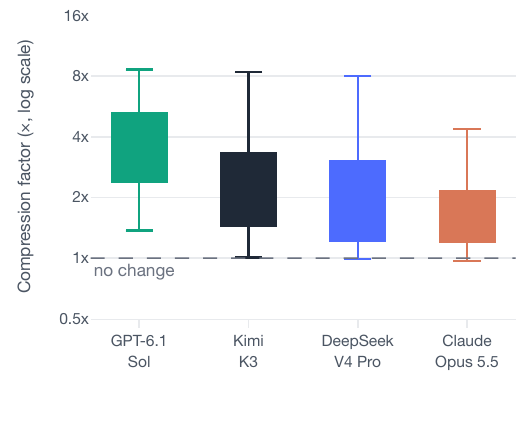}
  \end{subfigure}\hfill
  \begin{subfigure}[t]{0.48\linewidth}
    \centering
    \paneltitle{Compactions later reopened}{\% of compaction calls, content retrieved again}
    \includegraphics[width=\linewidth]{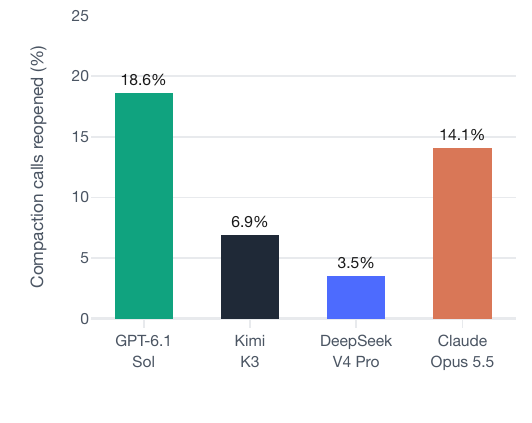}
  \end{subfigure}
  \caption{\textbf{Left:} compression factor per compaction call, the input tokens of the request in which the model decided to compact divided by the input tokens of the next request, which also carries the executed cell's output. A factor of $2\times$ means the context was halved; below $1\times$ means it grew. Boxes span the middle half of calls with the median inside, whiskers span the 5th to 95th percentile, and the axis is logarithmic. All confirmed compaction calls are pooled across context budgets. \textbf{Right:} share of compaction calls whose archived events were later retrieved again (\inlinecode{self.events[...]} or \inlinecode{.get}). This is a lower bound.}
  \label{fig:collapse}
\end{figure}

The effect of each compaction varies as well. \textbf{When GPT compacts its context, the request shrinks by a median of $3.7\times$ (Figure~\ref{fig:collapse}, left).} The other models shrink it less: the median is $2.1\times$ for Kimi, $1.8\times$ for DeepSeek, and $1.5\times$ for Opus. Compacting harder is not free, though. Compacted events are archived, and models sometimes go back for them: \textbf{GPT retrieves the archived content again after 19\% of its compaction calls, against 7\% for Kimi and 3\% for DeepSeek (Figure~\ref{fig:collapse}, right).} Opus is at 14\%, but it compacts so rarely that this estimate is noisy. Taken together, GPT looks like the model most comfortable with the context API: it knows when to call it, compacts the most per call, and retrieves archived content when it needs it.

\begin{figure}[tbp]
  \centering
  \figlegend{%
    \legitem{\legbox{nvamber}}{Return an answer}%
    \legitem{\legbox{nvgreen}}{Compact straight away}%
    \legitem{\legbox{nvblue}}{Read docs, then compact}%
    \legitem{\legbox{nvpurple}}{Inspect history, then compact}%
    \legitem{\legbox{nvlightgray}}{Other}}
  \includegraphics[width=0.245\linewidth]{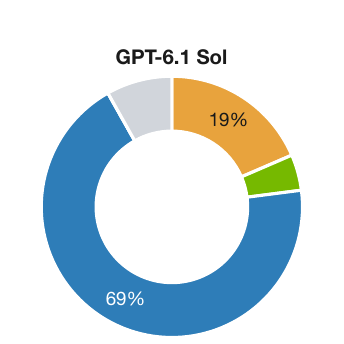}\hfill
  \includegraphics[width=0.245\linewidth]{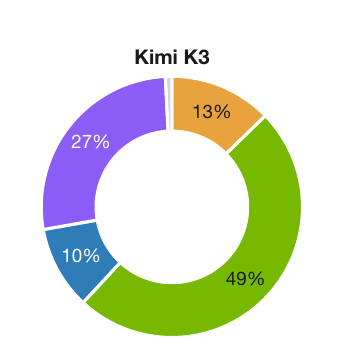}\hfill
  \includegraphics[width=0.245\linewidth]{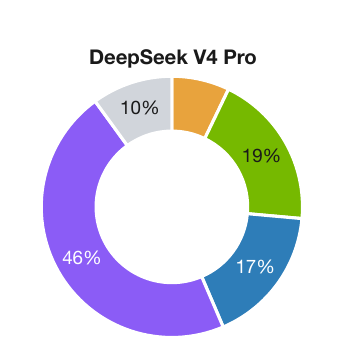}\hfill
  \includegraphics[width=0.245\linewidth]{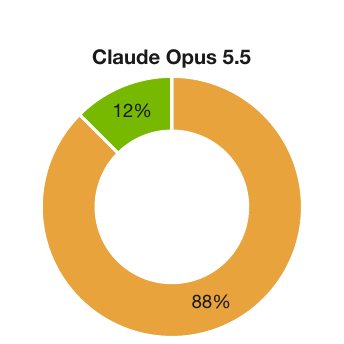}
  \caption{\textbf{What models do under context pressure.} Share of first-warning episodes by how they resolved, 64K context budget. When the context approaches its limit, we send the model a warning and observe how it responds.}
  \label{fig:pressure}
\end{figure}

Models also differ in what they do when the context fills up. At a 64K budget, we followed each model from the first context-pressure warning until the episode ended, either with an accepted final answer or with a compaction call. Opus mostly acknowledges the warning and wraps up: it returns an answer in most of its episodes instead of continuing. The other models mostly compact, and many consult something first: GPT reads the API docs before compacting in 69\% of episodes and DeepSeek inspects the event history first in 46\%, while Kimi compacts straight away in 49\% (Figure~\ref{fig:pressure}). This is consistent with Opus calling the API so rarely in Figure~\ref{fig:adoption}.

\textbf{This makes training a natural question.} Working in a harness that compacts history automatically does not necessarily teach a model when to compact, what to remove, or how to keep useful information.

Harnesses such as Mainplane~\citep{mainplane} now offer context management similar to \nooa's, and Codex's experimental mode with history, notes, and a new-context tool~\citep{codexpr} makes it plausible that related capabilities receive attention during GPT development. We conjecture that training for model-driven context management may contribute to GPT's stronger results. The feature's existence does not tell us how the evaluated model was trained, however, or whether any such training transfers to \nooa's interface.

\textbf{Our results motivate investigating how training can make context management more reliable and effective.}

\section{Context Anxiety}
\label{sec:anxiety}

Even models that use the API can compact more aggressively than they need to, keeping their context small while space remains. We call this \textbf{context anxiety}, following prior work~\citep{igbinedion2026lost} on models that misjudge their remaining token budget.

\begin{figure}[tbp]
  \centering
  \figlegend{\legitem{\legline{nvgreen}}{model-driven (85\% instruction)}\legitem{\legline{nvamber}}{no timing instruction}}
  \includegraphics[width=0.49\linewidth]{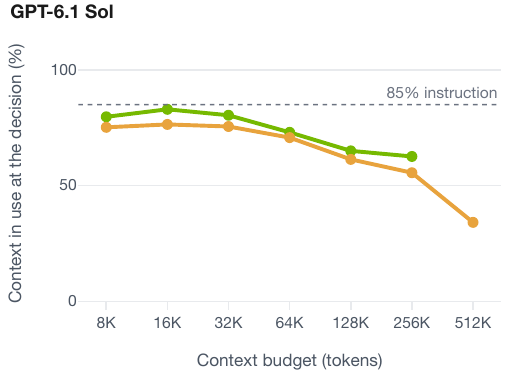}\hfill
  \includegraphics[width=0.49\linewidth]{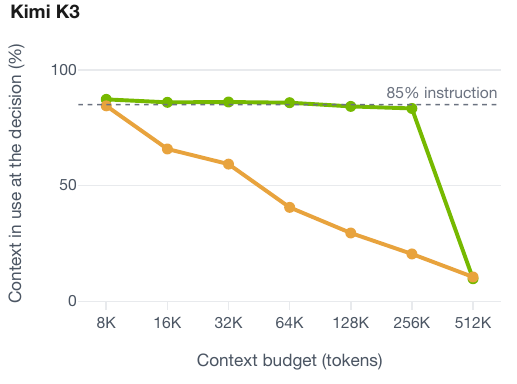}\\[0.4em]
  \includegraphics[width=0.49\linewidth]{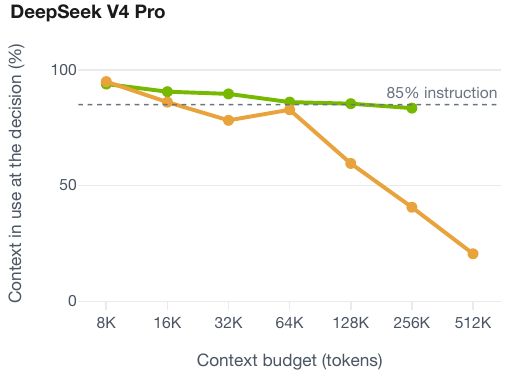}\hfill
  \includegraphics[width=0.49\linewidth]{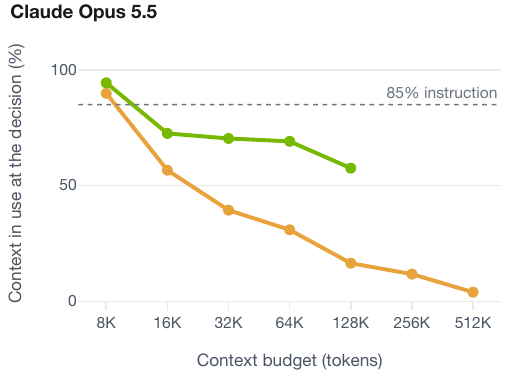}
  \caption{\textbf{Context in use at the first compaction:} median \% of the context budget, with and without the timing instruction. Each point is the median over tasks of the first compaction decision in that task. It measures when a model starts compacting. A budget with no point means the model never compacted at that budget.}
  \label{fig:anxiety}
\end{figure}

We measure compaction timing as \emph{the fraction of the context budget in use when the model makes its first compaction call in a task}. With the instruction to consider compacting only at 85\% usage, most models wait until they are close to that point. \textbf{Removing the instruction changes the behavior substantially for some of them (Figure~\ref{fig:anxiety}).}

Kimi and DeepSeek compact much earlier without the instruction. Kimi's median first compaction drops from about 84\% of the budget to 29\% at 128K, and DeepSeek's from 83\% to 40\% at 256K. GPT is less sensitive, which may reflect how it was trained. Opus almost never compacts with the instruction, but without it, it compacts regularly and early, at a median of 16\% of the budget at 128K.

\begin{figure}[tbp]
  \centering
  \figlegend{\legitem{\legdash{harnessgray}}{harness-driven}\legitem{\legline{nvgreen}}{model-driven}}
  \includegraphics[width=0.72\linewidth]{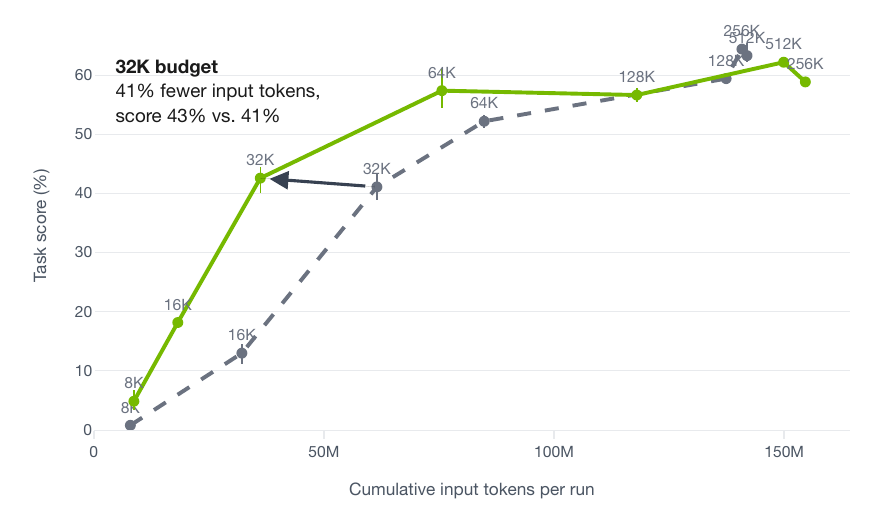}
  \caption{\textbf{Kimi K3: score vs.\ cumulative input tokens} on OfficeQA Pro v2, each point labelled with its context budget. Cumulative input counts all input tokens, cached or not, at every turn.}
  \label{fig:kimi-input}
\end{figure}

Kimi K3 shows the effect in token usage too (Figure~\ref{fig:kimi-input}). At a 32K budget, both strategies have the same maximum context size, yet the model-driven agent processes \textbf{41\% fewer cumulative input tokens (36M vs.\ 62M)} and scores slightly higher (43\% vs.\ 41\%). The same budget does not mean the same use of it.

Keeping less context can come at a cost. At budgets of 128K and above, the variants without the timing instruction score about 8 points below the harness baseline for Kimi and DeepSeek. GPT and Opus show no comparable change in accuracy.

Why do some models compact early? Many models are trained to keep their context short. Kimi K3, for example, is trained with RL penalties for overly long rollouts~\citep{kimik3}, which may teach it to economize on tokens even when space remains.

\section{From Fewer Tokens to Cheaper Inference}
\label{sec:cheaper}

\begin{figure}[tbp]
  \centering
  \figlegend{\legitem{\legdash{harnessgray}}{harness-driven}\legitem{\legline{nvgreen}}{model-driven}}
    \includegraphics[width=0.49\linewidth]{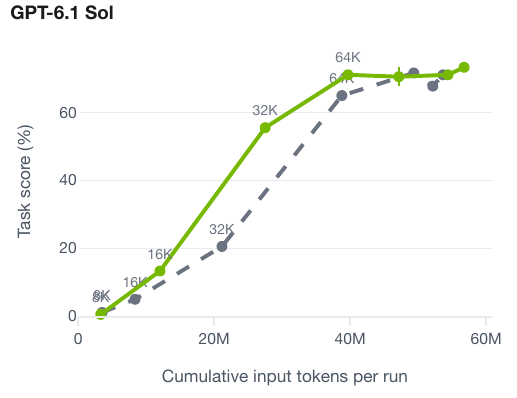}\hfill
    \includegraphics[width=0.49\linewidth]{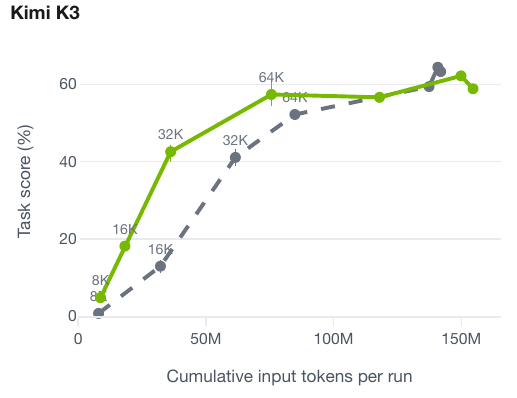}\\[0.4em]
    \includegraphics[width=0.49\linewidth]{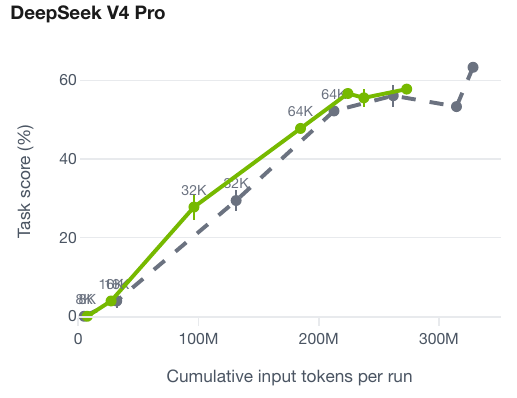}\hfill
    \includegraphics[width=0.49\linewidth]{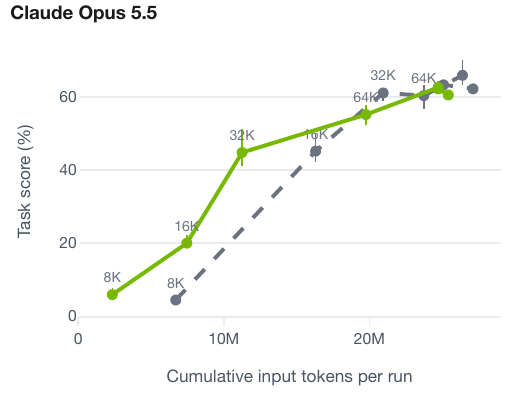}
  \caption{\textbf{Score vs.\ token usage, OfficeQA Pro v2}, harness-driven vs.\ model-driven, by model. Cumulative input tokens; points at 8K--64K are labelled with their context budget. Cumulative input counts all input tokens, cached or not, at every turn. Points are ordered by tokens, so labels may not ascend. Runs below 95\% trace coverage are excluded.}
  \label{fig:score-input}
\end{figure}

\begin{figure}[tbp]
  \centering
  \figlegend{\legitem{\legdash{harnessgray}}{harness-driven}\legitem{\legline{nvgreen}}{model-driven}}
    \includegraphics[width=0.49\linewidth]{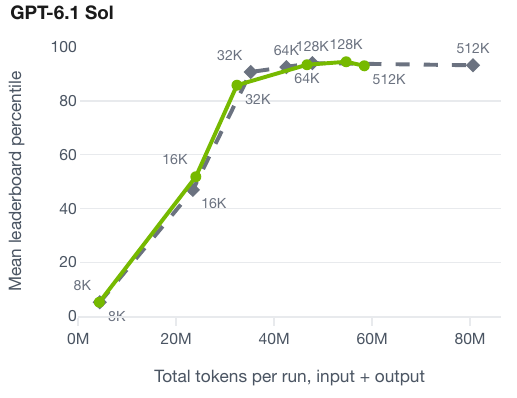}\hfill
    \includegraphics[width=0.49\linewidth]{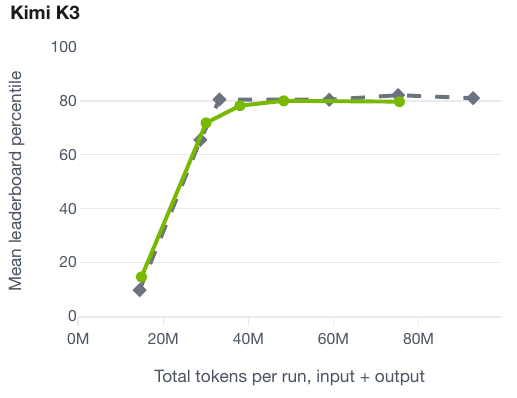}\\[0.4em]
    \includegraphics[width=0.49\linewidth]{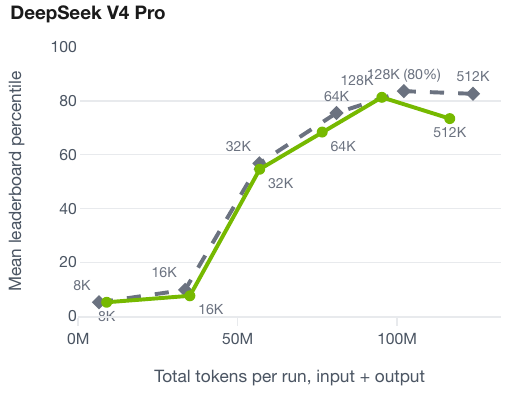}\hfill
    \includegraphics[width=0.49\linewidth]{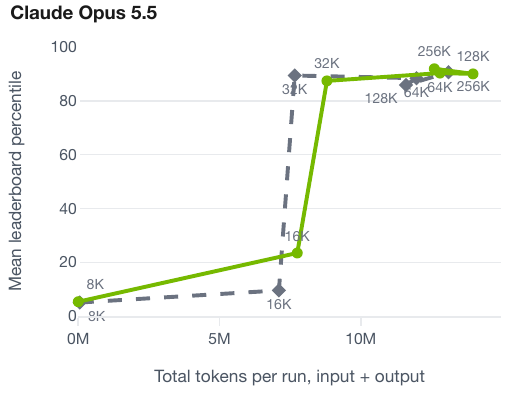}
  \caption{\textbf{Score vs.\ token usage, MLE-bench Lite}, harness-driven vs.\ model-driven, by model. Total input + output tokens, points labelled with their context budget. Token usage is the total recorded input and output tokens per run, including repeated prompts. Lines join increasing budgets and may turn back in token usage. Kimi K3 points combine several context budgets using equal-weight averages.}
  \label{fig:score-input-mle}
\end{figure}

Context management pays off most when a model reaches equal or better accuracy with fewer input tokens. Model-driven context management can do exactly that: \textbf{at a 64K budget, Kimi uses about 11\% fewer cumulative input tokens than its harness baseline while scoring higher (57\% vs.\ 52\%)}. The effect varies by model: GPT shows no comparable saving, and the other models fall in between (Figures~\ref{fig:score-input} and~\ref{fig:score-input-mle}).

\begin{figure}[tbp]
  \centering
\begin{tikzpicture}[x=0.85\linewidth/720, y=-0.85\linewidth/720, every node/.style={inner sep=0pt, outer sep=0pt}]
  \tikzset{
    b/.style={line width=0.5pt, rounded corners=1.5pt},
    cache/.style={b, fill=diaggreenfill, draw=nvgreen},
    fresh/.style={b, fill=diagamberfill, draw=nvamber},
    row/.style={anchor=base west, font=\sffamily\scriptsize, text=diagmute},
    note/.style={anchor=base west, font=\sffamily\scriptsize, text=diagmute},
    notec/.style={note, anchor=base},
  }
  \newcommand{\blk}[6]{\draw[#5] (#1,#2) rectangle ++(#3,#4);
    \node[font=\sffamily\scriptsize, text=diagink] at ({#1+#3/2},{#2+#4/2}) {#6};}

  \node[row] at (0,42) {Before edit};
  \blk{120}{20}{180}{36}{cache}{A $\cdot$ cached}
  \blk{304}{20}{180}{36}{cache}{B $\cdot$ cached}
  \blk{488}{20}{180}{36}{cache}{C $\cdot$ cached}

  \node[row] at (0,116) {After compacting B};
  \blk{120}{94}{180}{36}{cache}{A $\cdot$ still cached}
  \blk{304}{94}{56}{36}{fresh}{\tiny Summary}
  \blk{364}{94}{180}{36}{fresh}{C $\cdot$ recomputed}

  \draw[nvamber, line width=0.8pt] (304,140) -- ++(0,6) -- ++(240,0) -- ++(0,-6);
  \node[note] at (304,164) {fresh computation: summary + everything after the edit};
  \draw[nvamber, line width=0.8pt] (120,140) -- ++(0,6) -- ++(180,0) -- ++(0,-6);
  \node[notec] at (210,164) {prefix reused};

  \blk{120}{196}{18}{18}{cache}{}
  \node[note] at (146,209) {reused from KV cache};
  \blk{320}{196}{18}{18}{fresh}{}
  \node[note] at (346,209) {needs fresh computation};
\end{tikzpicture}
  \caption{Compacting B shortens the prompt, but the summary and everything after it must be computed again.}
  \label{fig:cache}
\end{figure}
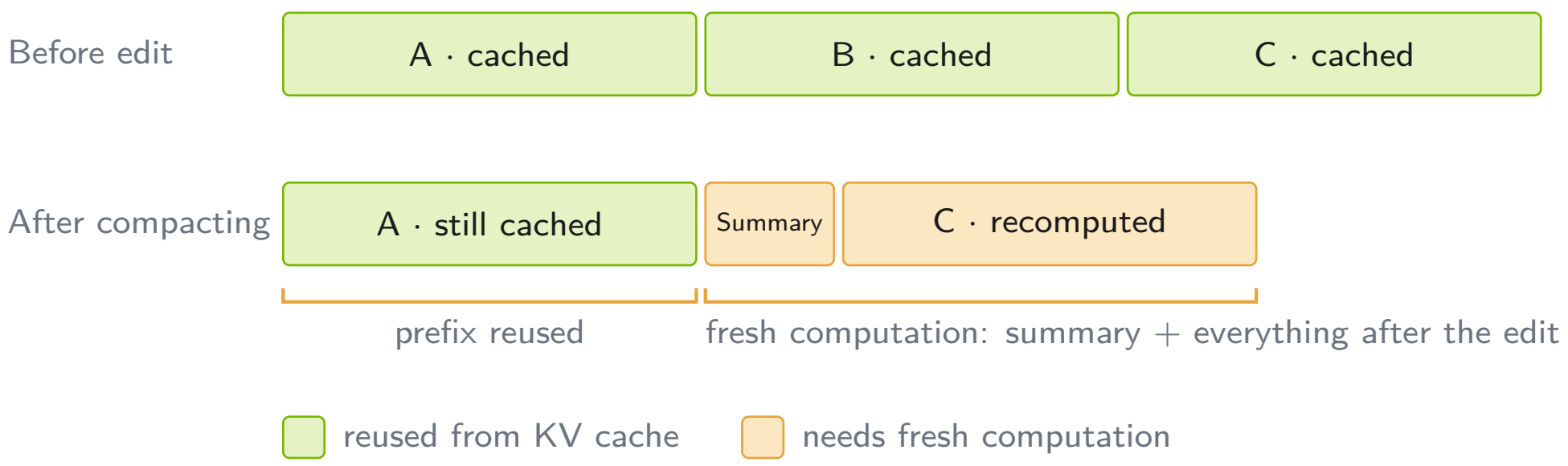

However, input-token counts tell only part of the story. Inference engines avoid reprocessing the prompt on every turn by reusing cached computation (the KV cache) for the prefix of the prompt that hasn't changed~\citep{zheng2024sglang,vllmprefixcaching}. Editing the context, for example replacing part of the history with a summary, keeps that cache valid up to the edit but invalidates everything after it, even text that didn't change (Figure~\ref{fig:cache}), so a shorter context isn't automatically cheaper. It also depends on how much the engine has to recompute after each edit.

\begin{figure}[tbp]
  \centering
  \figlegend{\legitem{\legbox{diagink}}{Model-driven (\nooa's context API)}\legitem{\legbox{diagink!40}}{Harness-driven (summarize, keep last 10 events)}}
  \begin{subfigure}[t]{0.54\linewidth}
    \centering
    \paneltitle{Reduction in prefill tokens per request}{\% vs.\ standard prefix caching, higher is better}
    \includegraphics[width=\linewidth]{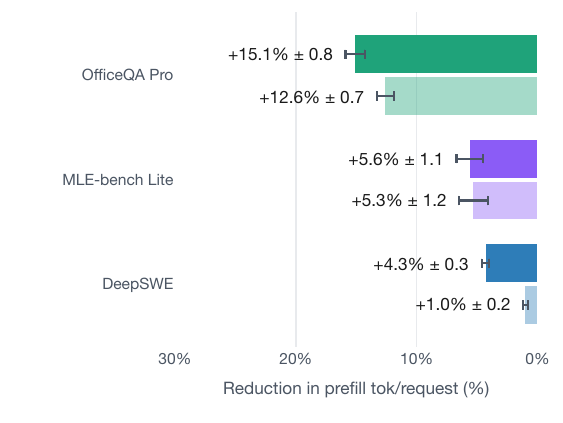}
  \end{subfigure}\hfill
  \begin{subfigure}[t]{0.43\linewidth}
    \centering
    \paneltitle{Task reward delta}{suffix KV reuse minus no KV reuse, with $\pm$ error}
    \includegraphics[width=\linewidth]{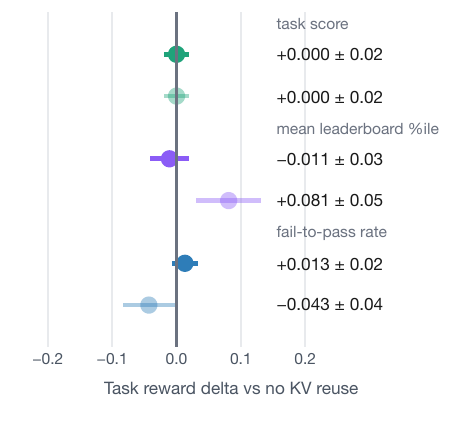}
  \end{subfigure}
  \caption{Suffix KV reuse versus standard prefix caching, Qwen3.6-27B at 32K context, under model-driven (\nooa's context API) and harness-driven (summarize, keep the last 10 events) context management. \textbf{Left:} reduction in prefilled input tokens per request. \textbf{Right:} change in reward, measured in each benchmark's own metric (DeepSWE: fail-to-pass rate; MLE-bench Lite: mean leaderboard percentile; OfficeQA Pro: task score).}
  \label{fig:suffix}
\end{figure}

To address this, recent papers such as SideQuest~\citep{kariyappa2026sidequest} from NVIDIA, Context Language Models~\citep{shao2026clm}, and Cache Engineering for Agents~\citep{fu2026cache} propose to reuse the KV cache of unchanged \emph{suffix} text after an edit instead of recomputing it, which is an approximation aimed at preserving useful model state while reducing serving compute work. We paired \nooa's context API with a suffix KV reuse prototype implemented in the SGLang serving engine~\citep{zheng2024sglang}, using Qwen3.6-27B~\citep{qwen2026qwen36}. On MLE-bench Lite, OfficeQA Pro and the agentic coding benchmark DeepSWE~\citep{deepswe}, it \textbf{cut prefill compute by 4--15\% per request compared with standard prefix caching, with no measurable impact on task accuracy} (Figure~\ref{fig:suffix}). Savings are largest on OfficeQA Pro because the suffix text that survives an edit is large, about 10K tokens or roughly 60\% of the prompt on average, because the agent's tool calls retrieve long documents that remain in context.

\textbf{These results suggest that model-driven context management opens up a new design space for agentic systems.} When models actively shape their own context, the harness and the inference engine no longer just carry out their own roles. They become partners in deciding what to keep and how to serve it efficiently. We are actively exploring this co-design of model, harness and inference engine to push the accuracy-efficiency frontier for agentic workloads.

\section{Limitations}
\label{sec:limitations}

Our results should be read with the following caveats.

\begin{itemize}[leftmargin=1.5em, itemsep=0.3em]
  \item \textbf{Small benchmarks.} OfficeQA Pro v2 has 90 questions and MLE-bench Lite has 22 competitions, so small differences between strategies or models may not be significant.
  \item \textbf{Simulated context pressure.} We shrink the context budget to emulate longer tasks. This is a proxy: it does not capture every property of genuinely long-horizon work, and the fallback that drops the oldest events at the limit can itself affect scores.
  \item \textbf{Four models, unknown training.} We evaluate four models and do not know how they were trained. Our explanation for GPT's stronger results is a conjecture, not a tested claim.
  \item \textbf{Single open model for KV reuse.} The suffix KV-reuse prototype was evaluated with one open model (Qwen3.6-27B) in one serving engine. It is an approximation to full recomputation, and we measured prefill compute, not end-to-end latency or cost.
\end{itemize}

\section{Related Work}
\textbf{Models that manage their own context.}
A growing line of work gives models explicit control over the information
available to them. MemGPT~\citep{packer2023memgpt} introduces hierarchical
memory management, while Recursive Language Models~\citep{zhang2025rlm}
place long inputs in an external programming environment that the model
can inspect and process through recursive calls. More directly concerned
with the evolving reasoning or interaction context,
MEMENTO~\citep{kontonis2026memento},
AgentFold~\citep{ye2025agentfold}, and
Context-Folding~\citep{sun2025contextfolding}
develop learned mechanisms for retaining, condensing, or restructuring
past computation.
SelfCompact~\citep{li2026selfcompacting} studies model-triggered
compaction with a prompting rubric and no fine-tuning.
Context Language Models~\citep{shao2026clm} expose the live context as
an editable file, evaluating both existing models and learned
context-management strategies, while
SideQuest~\citep{kariyappa2026sidequest} uses an auxiliary reasoning
task to select KV cache content for removal.
Our work provides a complementary empirical study of current models
under a shared context-management API, focusing on how their behavior
and task performance change with context pressure and how these
changes interact with serving efficiency.

\textbf{Efficient serving of edited contexts.}
Standard prefix caching reuses cached states only for a matching
prefix, so edits within the history can require re-prefilling the
retained suffix. For agent context management,
Context Language Models~\citep{shao2026clm} introduce Suffix Cache
Reuse to retain cached states beyond the matching prefix.
A related approach operates directly on the active KV cache:
SideQuest~\citep{kariyappa2026sidequest} removes selected cache content,
and Cache Engineering for Agents~\citep{fu2026cache} co-designs cache
operations and a serving engine to avoid re-prefilling retained
context. We pair a prototype supporting suffix cache reuse with a
model-driven context API and measure its effect on prefill compute
and accuracy (Section~\ref{sec:cheaper}).
\section{Discussion and Conclusion}
\label{sec:next}

We compared harness-driven and model-driven context management across four models and two benchmarks. Giving models control over their own context can improve accuracy and often reduces input tokens, especially at tight budgets, but the benefit depends heavily on the model: models differ in how often they use the API, how aggressively they compact, and when they start. Inference engines can make the resulting edits cheaper by reusing the KV cache of the unchanged suffix.

We expect model-driven context management to become a standard capability of frontier models. Getting there takes two things: training that teaches models when and what to compact, and inference engines that make context edits cheap. This points to two directions:

\begin{itemize}[leftmargin=1.5em, itemsep=0.3em]
  \item \textbf{Improve the underlying capability.} Some models already use context APIs they were never trained on, and GPT gains substantially from it at tight budgets. We believe training models to use these APIs correctly and effectively will make the capability reliable across models.
  \item \textbf{Connect the harness to the inference engine.} Design serving so that context edits preserve reusable computation and release KV cache entries that are no longer needed.
\end{itemize}

\FloatBarrier
\bibliographystyle{plainnat}
\bibliography{references}

\end{document}